\documentclass[runningheads]{llncs}

\usepackage{amsmath}
\usepackage{algorithm}
\usepackage[noend]{algpseudocode}
\usepackage{nicefrac}
\usepackage{cleveref}
\usepackage{booktabs}
\usepackage{bbding}

\DeclareMathOperator*{\argmin}{arg\,min}
\algnewcommand\AND{\quad \textbf{AND}\quad}
\algnewcommand\OR{\quad \textbf{OR}\quad}

\usepackage[T1]{fontenc}
\usepackage{graphicx}
\begin{document}
\title{Constrained Machine Learning Through Hyperspherical Representation}
%
%
\author{Gaetano Signorelli \Envelope \orcidID{0009-0005-5221-0717} \and Michele Lombardi \orcidID{0000-0003-4709-8888}}
\authorrunning{G. Signorelli, M.Lombardi}
%
\institute{University of Bologna\\
\email{\{gaetano.signorelli2, michele.lombardi2\}@unibo.it}}
\maketitle              
\begin{abstract}
The problem of ensuring constraints satisfaction on the output of machine learning models is critical for many applications, especially in safety-critical domains.
Modern approaches rely on penalty-based methods at training time, which do not guarantee to avoid constraints violations; or constraint-specific model architectures (e.g., for monotonocity); or on output projection, which requires to solve an optimization problem that might be computationally demanding.
We present the Hypersherical Constrained Representation, a novel method to enforce constraints in the output space for convex and bounded feasibility regions (generalizable to star domains). Our method operates on a different representation system, where Euclidean coordinates are converted into hyperspherical coordinates relative to the constrained region, which can only inherently represent feasible points.
Experiments on a synthetic and a real-world dataset show that our method has predictive performance comparable to the other approaches, can guarantee $100\%$ constraint satisfaction, and has a minimal computational cost at inference time.


\end{abstract}

\section{Introduction}

Modern machine learning (ML) techniques are widespread in a variety of fields, including some in which the predicted output must satisfy a set of constraints, for consistency (e.g., physical laws) or safety (e.g., autonomous driving) reasons.

State-of-the-art methods are generally based on architectural choices, output projection or penalty-based approaches. 
Architectural choices can be applied only for specific constraints, satisfied by design by selecting appropriate ML models; for instance, the architectures from \cite{gupta2016monotonic}, \cite{you2017deep} and \cite{sivaraman2020counterexample} can specifically handle monotonicity constraints.
Output projection avoids constraint violation by identifying, given a possibly infeasible prediction, the feasible point that has either minimum distance or maximum likelihood.
However, this operation requires solving a possibly costly optimization problem. 
Penalty-based approaches introduce extra terms in the loss function (see e.g. \cite{xu2018semantic}) that penalize constraint violation.
These methods incur no inference-time overhead and, under certain assumptions, can guarantee constraint satisfaction.
However, guarantees are limited to the training data and do not apply to unseen examples.

In this paper, we propose the \textit{Hyperspherical Constrained Representation}: a novel solution, which can provide feasibility guarantees both in and out of the training distribution, for convex and bounded constrained spaces (and generalizable to star domains with some careful design choices).
The method relies on a conversion from the canonical Euclidean space to an alternative system inspired by hyperspherical coordinates, designed to be incapable of representing infeasible points.
The conversion introduces minimal overhead at inference time and enables training via classical supervised learning.

Our main contribution is the introduction of an original method to enforce constraints in ML for convex and bounded regions that is: (i) capable of providing satisfaction guarantees; (ii) significantly faster than projection-based strategies; and (iii) competitive in terms of accuracy with the alternative approaches.

\section{Problem statement}

We focus on a specific class of hard constraints in the output space of a ML model, i.e. those whose conjunction defines a convex and bounded region.
The method can also be generalized to star domains (i.e. sets where every point can be connected by a line to a single origin) with some careful design choices.

Formally, let $\mathcal{X} \subseteq \mathcal{R}^k$ and $\mathcal{Y} \subseteq \mathcal{R}^n$ be respectively the input and output space for a ML task;
let $\{x_j, y_j\}_{j=1}^N$ be the training set and let $f_{\theta}: \mathcal{X} \rightarrow \mathcal{Y}$ be a function (predictive model) parameterized on $\theta$.
Let $\mathcal{L}: \mathcal{Y} \times \mathcal{Y} \rightarrow \mathcal{R}$ be a loss function. Given a set of $m$ constraints $c_1(y), c_2(y), \dots, c_m(y)$ defined as convex functions on the output space $\mathcal{Y}$, let:
\begin{equation}
    C(y) = \bigwedge_{i=1}^m (c_i(y) \leq 0)
\end{equation}
be the predicate defining the feasible region, which is convex due to the convexity of the $c_i(y)$ functions.
Then the problem of learning a model with feasibility guarantees can be formalized as:
\begin{equation}
\argmin_{\theta} \left\{
    \sum_{j=1}^N
    \mathcal{L}(y_j, f_\theta(x_j))
    \text{ s.t. }
    C(f_\theta(x)) \quad \forall x \in \mathcal{X}
    \right\}
\end{equation}
where the loss is computed on the training set, but the constraints should hold for every point in the input space.

\section{Related work}

A typical approach to enforce constraints in ML models consists in adding penalty terms to the loss function at training time.
The extra terms are weighted by $\lambda$ parameters, which under certain circumstances correspond to Langrangian multipliers \cite{rockafellar1993lagrange}.
The approach is usually presented as a form of \textit{regularization}.
Examples include the regularization technique from \cite{xu2018semantic}, which is based on the weighted model count presented by \cite{chavira2008probabilistic}.
In \cite{diligenti2017semantic,serafini2016logic} the penalty terms are derived 
via fuzzy logic from first-order logic constraints.
There are many other regularization approaches in the literature, e.g. \cite{fischer2019dl2} and \cite{hu2016harnessing}.
Regularization approaches can handle many types of constraints, but they cannot generally guarantee feasibility out of the training distribution.


A second common strategy to enforce constraints consists in projecting the model output $y$ to a feasible point $\hat{y}$ such that the property $p(\hat{y})$ holds.
This approach provides feasibility guarantees even on unseen examples, but it involves solving an optimization problem that may be computationally expensive.

A third viable option is to enforce constraints by designing specific ML models and training algorithms. These approaches tend to be restricted to specific constraints classes.
Examples include monotonic lattices \cite{gupta2016monotonic}, deep lattice networks \cite{you2017deep}, and COMET \cite{sivaraman2020counterexample}, which enforce monotonicity in neural networks; specific classes of safety related constraints are considered in \cite{leino2022self,lin2020art}.
The Multiplexnet from \cite{hoernle2022multiplexnet} is a much more general example, where a layer is responsible for satisfying constraints in disjunctive normal form (DNF), provided that the individual terms of the disjunction are sufficiently simple.
ProbLog \cite{manhaeve2021neural} exploits its own framework to train neural networks that satisfy constraints, but only discrete variables can be modeled.
The DeepSaDe method from \cite{goyal2024deepsade} relies on a custom network architecture and a training algorithm that solves a Max SMT model at training time, this can guarantee feasibility for a broad class of constraints, but at the cost of a very long training time and a loss in accuracy.

Compared to the previous methods, our approach guarantees constraint satisfaction in and out of distribution with no overhead at inference time, has accuracy on par with the best alternatives, and can be trained via classical supervised learning after a pre-processing step (with small or limited computational cost).


\section{Method}

We now present the Hyperspherical Constrained Representation (HCR), which exploits a change in representation to enforce constraints.
In detail, we apply a transformation in the output space from the canonical Euclidean representation to a coordinate system that spans exactly the feasible region.
%
%
We propose a representation system inspired by the hyperspherical coordinates, where each point is represented by two elements: an angle and a distance with respect to an origin.
Similarly, our hyperspherical system relies on fixing an origin, which has to be feasible, and then representing points in the output space by means of a direction, expressed as a normalized vector, and a distance, expressed as a value in the range $(0, 1)$.
This pair determines the relative position of the point along the segment connecting the origin with the frontier of the feasible region.
Convexity ensures that every distance value identifies a feasible point, boundedness ensures that a unique point on the frontier is associated to a given direction.
No point outside of the frontier can be represented by construction and, since the direction vector is unrestricted, any feasible point can be represented in this fashion.
HCR works by predicting directly into the hyperspherical space and then converting back to Euclidean coordinates.

\paragraph{Formalization:}

Let $O \in \mathcal{R}^n \mid C(O)$ be the origin of the system. Each point $y \in \mathcal{R}^n \mid C(y)$ can be converted to hyperspherical coordinates using a function:
\begin{equation}
    \phi(y): \mathcal{R}^n \rightarrow \mathcal{R}^n \times [0, 1]
\end{equation}
Let $\phi(y) = (d, r)$, such that $d \in \mathcal{R}^n$ is the point direction and $r \in [0, 1]$ is the point distance; these are defined as:
\begin{equation}
    d(y) = \frac{y-O}{\|y-O\|_2}, \quad r(y) = \frac{\|y-O\|_2}{\mathrm{s}(d(y))}
\end{equation}
where $s$ determines the distance from $O$ and the intersection between the ray identified by $d(y)$ and the frontier of the feasible region. This is defined as:
\begin{equation} \label{eq: compute-s}
    s(d) = \min \{t \mid \exists i \in \{1, 2, \dots m\} \mid c_i(t\, d)=0\}
\end{equation}
Conversely, hyperspherical coordinates can be converted into Euclidean ones using the inverse of $\phi$:
\begin{equation}
    \phi(y)^{-1}: \mathcal{R}^n \times [0, 1] \rightarrow \mathcal{R}^n
\end{equation}
which is defined as:
\begin{equation} \label{eq: inverse}
    y = O + d \cdot r \cdot s(d)
\end{equation}
The bottleneck in both conversion processes is computing $s(d)$, which requires in the worst case to solve $m$ equations in the form $c_i(t\, d) = 0$ in the scalar variable $t$.
For many constraint types, this can be done efficiently via the Brent's, Newton-Raphson's or Halley's method.
Notably, since computing intersection is typically less expensive than solving a constrained optimization problem, our method can be expected to be faster than projection at inference time.

\paragraph{A Numerical Example:}

\begin{figure}[tb]
    \centering
    \includegraphics[width=0.6\textwidth]{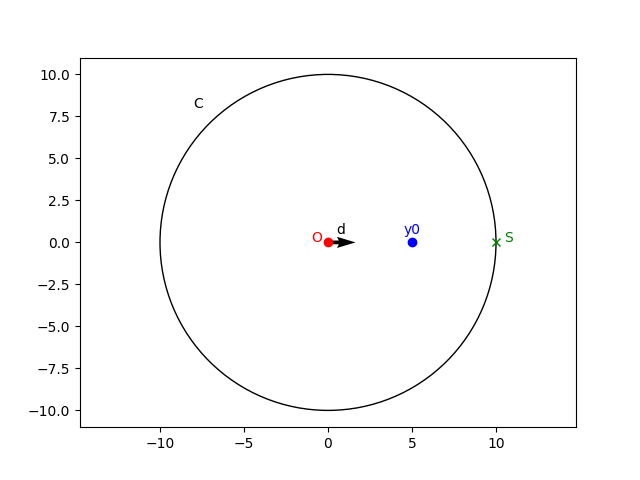}
    \caption{Example of an instance of conversion. The constrained region is $C$: a circle centered at $O=(0,0)$ and radius $R=10$. The point $y_0=(5, 0)$ is converted to hyperspherical coordinates $(d=(1, 0), r=0.5)$ by finding the intersection point $S$.}
    \label{fig:example}
\end{figure}

To better explain our method, we include a simple example of the conversion procedure. Let us assume that $n=2$ and $m=1$, with the constrained region being a circular area of radius equal to $10$ and centered around the origin $O = (0, 0)$, i.e. $C \equiv \|y\|_2 - 10 \leq 0$.
%
%
Given a point $y_0 = (5, 0)$, the hyperspherical coordinates are computed by:
\begin{enumerate}
    \item Compute $d_0 = d(y_0) = \nicefrac{y_0}{5} = (1, 0)$.
    \item Compute $s_0 = s(d_0)$, corresponding to the minimum value $t$ such that $d_0  \cdot t_0$ intersects one constraint in $C$. In this example, there is a single constraint and the intersection along $d_0$ is trivially at $S = (10, 0)$, so that $s(d_0) = 10$. Formally, this is obtained by solving the equation $\|O + t \cdot d_0\|_2 - 10 = 0$.
    \item Compute $r_0 = r(y_0) = \nicefrac{5}{10} = 0.5$.
\end{enumerate}
Hence, the hyperspherical coordinates of $y_0$ are $((1, 0), 0.5)$. The objects involved in the example are all depicted in \cref{fig:example}.
Reconstructing $y_0$ involves computing the intersection point $S$ again and then applying \cref{eq: inverse}.
Note that, in practice, constraints with a fixed known radius $R$ with respect to the origin, as in this example, can be easily handled as $s=R$ by definition.

\paragraph{Acceleration Technique:}\label{sec:conversion}

In an effort to further speed up the solution of \cref{eq: compute-s}, we employ an acceleration technique.
Rather than computing the intersections with all $m$ constraints, we exploit the fact that any point sufficiently far from the original one along the direction $d$ will: (i) violate at least one constraint; (ii) typically violate only a subset of the constraints.
Based on this observation, we pick up a point along $d$ and gradually move it far away, via multiplication by a scaling hyperparameter (\texttt{BASE-MULTIPLIER}).
For a proper choice of the multiplier, a violation will be found in a few iterations; every iteration can be very fast, since checking constraint violation is much faster than computing an intersection.
Our experiments show that setting the \texttt{BASE-MULTIPLIER} to approximately the radial length of the constrained region reduces the search to $1-2$ constraints on average.
This technique is described in \cref{alg:restrict}, which computes and returns the restricted set of candidate constraints to be checked, in order to identify the intersection point.

\paragraph{Training Process:}

The training process requires to convert all points in the training set to their corresponding hyperspherical coordinates by means of $\phi$.
Then, the model can be trained via supervised learning on the hyperspherical space.
In case some training points are outside the feasible region, they need to be made feasible (e.g. via projection) before the hyperspherical conversion.
While the computational cost of this operation can be non negligible, it is only incurred at training time, where computational resources are typically plentiful and response times are not an issue.

\paragraph{Generalization to Star Domains:}

The main assumption for HCR requires $C$ to be a convex set, so that every line segment between two points lies inside the feasible region. This assumption allows to arbitrarily choose an origin, guaranteeing one and only one intersection point with the boundaries of $C$ along all the possible directions.
However, this property must hold only for the origin. Thus, the convexity assumption can be relaxed in favor of a star domain, assuming $C$ to be a radially convex set.
This relaxation expands the applicability of our method to all the cases where at least one point $s_0 \in C$ is known, such that $\forall s \in C$ the line segment between $s_0$ and $s$ lies in $C$.

\begin{algorithm}[tb]
    \caption{Restrict constraints trick}\label{alg:restrict}

    \begin{algorithmic}

      \Function{restrict-constraints}{$O, d$}
        \State $i \gets 0$
        \While{$i < \texttt{MAX-ITERATIONS}$}
            \State $mult \gets \texttt{BASE-MULTIPLIER} \cdot (1 + 0.5 i)$
            \State $point \gets O + d \cdot mult$
            \State $violated \gets \texttt{ENVIRONMENT.violated-constraints(point)}$
            \If{\texttt{length(violated}) > 0}
                \State \textbf{return} $violated$
            \EndIf
            \State $i \gets i + 1$
        \EndWhile
        \State \textbf{return} $\texttt{ENVIRONMENT}.constraints$
      \EndFunction
      
    \end{algorithmic}
\end{algorithm}

\section{Experiments}\label{sec:experiments}

We executed experiments on two distinct datasets, comparing our method with the following models: a simple neural network, with no constraint enforcement; a network trained with penalty (Lagrangian) terms, and multiplier calibrated via dual ascent \cite{fioretto2021lagrangian}; a network paired with inference-time projection.

\paragraph{Synthetic benchmark:}

We first built a synthetic dataset with a single constraint, i.e. a hypersphere in $n$ dimensions, centered at the origin and having a radius $R$.
We generate (feasible) data using different distributions for the training and test set, to reproduce common scenarios where violations may occur when performing out-of-distribution inference.
In detail, we sample $k$ features from uniform distributions, with $x_{\mathit{train}} \sim \mathcal{U}(-0.8, 0.8)$ and $x_{\mathit{test}} \sim \mathcal{U}(-1.0, 1.0)$.
We subsequently generate the weights matrix $W \in \mathcal{R}^n \times \mathcal{R}^k$ sampling from $\sim \mathcal{U}(-10.0, 10.0)$ and then normalizing so that $\sum_{i=1}^k W_{ij} = 1 \quad \forall j \in \{1, 2, \dots, n\}$. We generate $y_{\mathit{train}} = R \cdot Wx_{\mathit{train}}$ and $y_{\mathit{test}} = R \cdot Wx_{\mathit{test}}$. This procedure creates samples inside the hypercube circumscribed to the hypersphere. All the points outside of the feasible region are then projected inside it by solving: $\argmin_{\hat{y}} \|\hat{y} - y\|_2 \mid \|\hat{y}\|_2 < R$.
In all our experiments we have $R = 10$, $k=128$ and $n=768$, to stress both the computation cost and accuracy of the methods on complex scenarios. We use $500$ samples for training and $1000$ for testing.

\paragraph{M4 Forecasting Competition Dataset:}

As a second dataset, we use real time-series from the M4 forecasting competition \cite{MAKRIDAKIS202054}.
This dataset is made up of $100k$ time series, split in temporal categories (hourly, daily, weekly, quarterly and yearly).
The tasks consists in predicting values in an output window (with size $n$), based on observed values from an input window (with size $k$).
We impose a max-deviation constraint between consecutive points in the prediction window, i.e.
such that $|v_i - v_{i+1}| \leq d_{max} \forall i \in \{1, 2, \dots, n-1\}$.
The value $d_{max}$ corresponds to the largest deviation between two consecutive values in the training set.
Similarly, we impose an upper and lower bound for each value, so to define a convex constrained area in the form of a polytope in $\mathcal{R}^n$.
Our focus for these experiments is on violations that might occur from out-of-distribution inference, rather than from a lack of alignment between the constraints and the data distribution.
For this reason, we preprocess all time series via projection to ensure that the constraints are satisfied also in test data.
We use $20\%$ of the data as training set to encourage overfitting and constraint violations.
We set $k=n$, while $n$ is the same used in \cite{MAKRIDAKIS202054} and it depends on the temporal split.

\paragraph{Results:}

For both datasets, we implement a simple neural network consisting of an encoding layer and a linear regression head for all the models, except for ours, having two regression heads: one for $d$, followed by a normalization step; the other for $r$, followed by a sigmoid function. We train all the models using the Adam algorithm \cite{kingma2014adam} and the MSE loss function.
We use a single feed-forward layer for the synthetic case and a Long Short-term Memory model \cite{hochreiter1997long} for the M4 dataset as encoding layers.
We apply data standardization to target values for all the models except ours; $x$ is also standardized for the M4 dataset.
We repeated the experiments using $10$ different seeds for the synthetic dataset, and $30$ different time series for the M4 dataset. For this second dataset, we only report results based on the higher dimensional (most critical) temporal category (hourly), having $k=n=48$ and $m=190$. Results are shown in tables \ref{tab:results-synthetic} and \ref{tab:results-m4}.

\begin{table}[tb]
    \centering
    \begin{tabular}{lcccc}
    \toprule
      \textbf{Method} & \textbf{MSE} & \textbf{Inside ratio} & \textbf{Avg. time} & \textbf{Max time} \\
      \midrule
      Simple & $0.071 \pm 0.001$ & $0.341 \pm 0.009$ & \textit{NA} & \textit{NA} \\
      \hline
      Lagrangian & $0.068 \pm 0.018$ & $0.912 \pm 0.029$ & \textit{NA} & \textit{NA} \\
      \hline
      Projection & $0.050 \pm 0.001$ & $1.000 \pm 0.000$ & $0.0700 \pm 0.001$ & $0.1250 \pm 0.004$ \\
      \hline
      HCR & \boldmath {$0.010 \pm 0.002$} & $1.000 \pm 0.000$ & \boldmath $0.0001 \pm 0.000$ & \boldmath $0.0002 \pm 0.000$\\
      \bottomrule
    \end{tabular}
    \caption{Results on synthetic dataset. Mean-squared-error, percentage of feasible outputs, average and maximum post-processing time (projection or hyperspherical conversion) are reported for the test set.}
    \label{tab:results-synthetic}
\end{table}

\begin{table}[tb]
    \centering
    \begin{tabular}{lcccc}
    \toprule
      \textbf{Method} & \textbf{R-MSE} & \textbf{Inside ratio} & \textbf{Avg. time} & \textbf{Max time} \\
      \midrule
      Simple & $0.125 \pm 0.051$ & $0.894 \pm 0.111$ & \textit{NA} & \textit{NA} \\
      \hline
      Lagrangian & $0.191 \pm 0.050$ & $0.980 \pm 0.041$ & \textit{NA} & \textit{NA} \\
      \hline
      Projection & \boldmath $0.124 \pm 0.049$ & $1.000 \pm 0.000$ & $0.001 \pm 0.001$ & $0.004 \pm 0.011$ \\
      \hline
      HCR & $0.130 \pm 0.047$ & $1.000 \pm 0.000$ & \boldmath $0.0001 \pm 0.000$ & \boldmath $0.001 \pm 0.000$\\
      \bottomrule
    \end{tabular}
    \caption{Results on M4 dataset. Relative mean-squared-error, percentage of feasible outputs, average and maximum post-processing time (projection or hyperspherical conversion) are reported for the test set.}
    \label{tab:results-m4}
\end{table}

Experiments show similar performances in terms of MSE accuracy compared to the other methods; particularly for the synthetic dataset, where our method outperforms the baselines. Moreover, they highlight the gap in post-processing time between our method and projection: the HCR takes a negligible amount of time, with low variations depending on $n$, $m$ and the structure of the feasible region; projection takes a highly variable time, mostly depending on the nature of the given constraint, up to $700$ times higher than our approach for a single quadratic constraint in a high-dimensional space.

\section{Conclusions}

We have introduced a novel method to tackle constrained machine learning for convex and bounded feasible regions in output space. Our method guarantees constraints satisfaction using a conversion at inference time that requires, both theoretically and empirically, a negligible amount of time, outperforming projection methods, while still providing a comparable predictive accuracy.
We believe that the HCR method could be useful in settings such as control applications, where safe predictions are required and the limited resources make the use of projection difficult.
Furthermore, our method combined with a common root-finding method (e.g., Newton-Raphson) would be fully differentiable and it could be used in pipelines where constraints are required in intermediate steps (i.e., embeddings).
As a future work, we plan to further explore use cases for the discussed scenarios and directions to expand the method for non-convex regions.

\bibliography{references}

\end{document}